\documentclass{article}

\usepackage[dblblindworkshop, final]{neurips_2026}
\workshoptitle{NeurIPS 2026 Workshop on Principles of Generative Modeling (PriGM)}

\usepackage{hyperref}       
\usepackage{url}            
\usepackage{booktabs}       
\usepackage{amsfonts}       
\usepackage{nicefrac}       
\usepackage{microtype}      
\usepackage{xcolor}         
\usepackage{amsmath}
\usepackage{amssymb}
\usepackage{mathtools}
\usepackage{amsthm}
\usepackage[acronym]{glossaries}
\usepackage{microtype}
\usepackage{graphicx}
\usepackage{wrapfig}
\usepackage{algorithm}
\usepackage{algorithmic}
\usepackage{booktabs} 
\usepackage{cancel}
\usepackage{xcolor}
\usepackage{glossaries}
\usepackage{cleveref}
\usepackage{subcaption}
\theoremstyle{plain}

\theoremstyle{definition}

\theoremstyle{remark}

\renewcommand{\d}[1]{\ensuremath{\operatorname{d}\!{#1}}}

\newcommand{\defeq}{\stackrel{\text{\tiny def}}{=}}

\newcommand{\fp}{\overset{\rightarrow}{p}}

\newcommand{\expected}{\mathbb{E}}

\newcommand{\cumnoise}{\overline{\sigma}}

\newcommand{\diffusion}{\sigma}
\newcommand{\Eptot}{\mathcal{H}^{\text{tot}}}

\newcommand{\Ena}{\mathcal{H}^{\text{na}}}

\newcommand{\Eptotdot}{{\mathcal{H}}^{\text{tot}}}

\newacronym{ODE}{\textsc{ode}}{Ordinary Differential Equation}
\newacronym{CTMC}{\textsc{ctmc}}{Continuous Time Markov Chain}
\newacronym{KL}{\textsc{kl}}{Kullback-Leibler}
\newacronym{MLP}{\textsc{MLP}}{Multi-Layer-Perceptron}
\newacronym{NLP}{\textsc{NLP}}{Natural Language Processing}
\newacronym{GWAS}{\textsc{GWAS}}{Genome-Wide Association Study}
\newacronym{WSSS}{\textsc{WSSS}}{Weakly Supervised Semantic Segmentation}
\newacronym{CAM}{\textsc{CAM}}{Class Activation Map}
\newacronym{SDE}{\textsc{SDE}}{Stochastic Differential Equation}
\newacronym{DTC}{\textsc{DTC}}{Dual Total Correlation}
\newacronym{TC}{\textsc{TC}}{Total Correlation}
\newacronym{MI}{\textsc{MI}}{Mutual Information}
\newacronym{DWDSE}{\textsc{dwdse}}{Diffusion Weighted Denoising Score Entropy}
\newacronym{GP}{\textsc{GP}}{Gaussian Process}
\newacronym{CI}{\textsc{CI}}{Confidence Interval}
\newacronym{MEC}{\textsc{MEC}}{Minimum Entropy Coupling}

\newacronym{RND}{\textsc{RND}}{Radon-Nikodym Derivative}

\newacronym{FPK}{\textsc{FPK}}{Fokker Planck Kolmogorov}
\newacronym{VP}{\textsc{VP}}{Variance Preserving}

\newacronym{EDS}{\textsc{EDS}}{Entropic Discrete Schedule}
\newacronym{WDS}{\textsc{WDS}}{ Wasserstein Discrete Schedule}

\newacronym{UDM}{\textsc{UDM}}{Uniform Discrete Diffusion Model}

\newacronym{EP}{\textsc{EP}}{Entropy Production}

\newacronym{FID}{\textsc{FID}}{Frechet Inception Distance}

\newacronym{NFE}{\textsc{NFE}}{Number of Function Evaluations}

\title{Improved sampling schedules for Discrete Diffusion Models}

\author{%
  Alberto Foresti\thanks{Equal contribution.} \\
  EURECOM
  \And
  Mustapha Bounoua\footnotemark[1] \\
  EURECOM
  \AND
  Giulio Franzese \\
  EURECOM
  \And
  Luca Ambrogioni \\
  Donders Institute for Brain, Cognition and Behaviour \\ Radboud University
  \And
  Pietro Michiardi \\
  EURECOM
}

\begin{document}

\maketitle

\begin{abstract}
Uniform Discrete Diffusion Models have emerged as a powerful paradigm for generative modeling on sequence data; however, the information-theoretic principles governing their reverse processes remain underexplored. In this work, we bridge this gap by analyzing the reverse process dynamics through the lens of thermodynamic entropy production and dynamical activity. We use a variational expression of the discrete Wasserstein distance to characterize the dynamics of the generation process. Leveraging these insights, we first introduce the Entropic Discrete Schedule (EDS), which aligns with entropy production, and later construct the Wasserstein Discrete Schedule (WDS), which combines entropy production and dynamical activity to act as a geometric cost. Our experimental results on language modeling show that WDS provide the same results as uniform while using up to half of the computations on the MBPP benchmark. The code is available at \url{https://github.com/AlbertoForesti/impscheds}.  
\end{abstract}


\section{Introduction}

Discrete diffusion models \citep{austin2021structured, campbell2022continuous,lou2024discrete, sahoo2025simple, shi2024simplified} emerged as a powerful alternative to transformer-based approaches \citep{vaswani2017attention} due to innate parallel token generation capabilities and global context awareness unlocking advanced features such as non-linear infilling, bidirectional sequence refinement, and improved controllability \citep{schiff2024caduceus,nisonoff2024unlocking,li2024derivative}. With the emergence of large diffusion language models \citep{nie2025largelanguagediffusionmodels,xie2025dream}, research has focused on optimizing inference efficiency, exemplified by the fast solvers of \citet{ren2025fast}, as well as refining unmasking strategies for masked diffusion models \citep{sahoo2025simple,ben2025accelerated,kim2025train}. In particular, optimizing the sampling schedule of the reverse process has been identified as critical for efficiency \citep{chen2023importance}. On the other hand, recent work uncovered that \glspl{UDM} show potential in scaling better than their masked counterpart \citep{sahoo2026scaling, deschenaux2026diffusion, von2025scaling}, while naturally enabling self-correction. These results fundamentally challenge the view that masked diffusion is the future of generative modeling via discrete diffusion. Despite this, most schedule optimization strategies still focus on masked diffusion. Addressing this, \citet{park2024jump} analyzed the impact of compounding errors to propose a method for reducing generative costs, while \citet{fu2025learnable} introduced a distillation framework wherein a few-step student sampler learns to align its intermediate score trajectory with that of a high-quality teacher. While these advancements are valuable, the field currently lacks a fundamental understanding of the inner dynamics governing \glspl{UDM}. Specifically, it remains unclear how information flows, dissipates, and is reconstructed during the  discrete diffusion 
reverse process. A theoretical grounding in these dynamics is essential not only for interpreting model behavior but also for deriving improvements to sampling schedules. While diffusion models have been widely utilized to estimate information measures in both continuous \cite{franzese2023minde,kong2023information,bounoua2024somegai} and discrete settings \cite{foresti2026information}, the application of thermodynamic principles to analyze information storage \cite{premkumar2024neural} or optimize sampling strategies \cite{stancevic2025entropic} has been restricted to continuous diffusion models. 

\textbf{Contributions.} In this work, we build a physics-inspired framework for characterizing the dynamics of \glspl{UDM} with stochastic thermodynamics \citep{van2023thermodynamic}. We later use these quantities as physics-based heuristics for building sampling schedules for \glspl{UDM}. The schedule computation time for our schedule does not rely on an optimization process, and is therefore faster to compute. Our experimental results show that \gls{WDS} is robust across several tasks of increasing difficulty, providing up the same performance of the uniform schedule at up to half the number of \glspl{NFE} employed on the MBPP benchmark.

\section{Preliminaries}
\label{sec:preliminary}

Consider a \gls{CTMC} $X_t$, with $t \in[0,T]$, defined over a finite state space  $\mathcal{X} = \left\{1,\hdots,N\right\}$ and specified by the infinitesimal generators $Q_t$$: [0,T]\rightarrow \mathbb{R}^{N\times N}$, where the diagonal entries satisfy $Q_t(x,x) = -\sum_{y\neq x} Q_t(y,x),$, with $Q_t(x,y)\geq 0,\, x\neq y$. In this convention, \(Q_t(y,x)\) denotes the jump rate from the current state
\(x\) to the next state \(y\).The time evolution of the probability distribution $\text{Pr}(X_t=x)\defeq p_t(x)$, with \(p:=\{p_t\}_{[0,T]}\), satisfies the following \gls{ODE} \citep{anderson1991continuous}:

\begin{equation}\label{eq:forwardctmc}
\frac{d p_t}{dt} = Q_t\,p_t,
\qquad p_0 \approx p_{\mathrm{data}}.
\end{equation}
In this work, we focus on symmetric uniform-rate transition matrices \(Q_t(x,y)=\nicefrac{\sigma(t)}{N-1}\) and we assume that \(Q_t\) factorizes across dimensions such that \(Q_t(x,y)=0\) if the Hamming distance \(d(x,y)\) is larger than \(1\). Under appropriate conditions \citep{kelly1979reversibility}, the \gls{CTMC} can be reversed, allowing sampling from \(p_{\mathrm{data}}\) with the reverse score \(s_t(x)_y=\nicefrac{p_t(y)}{p_t(x)}\). In practice \(s_t(x)_y\) is generally intractable but can be estimated via a neural network \(s_t^\theta(x)_y\) \citep{lou2024discrete}.


\section{Methodology}

\label{sec:3}
The generative CTMC can be analyzed through stochastic
thermodynamics, where irreversibility of a process is quantified by the total entropy
production. At the trajectory level, this quantity measures the breaking of
time-reversal symmetry as the Kullback-Leibler divergence between the path law
of the process and that of its time-reversed counterpart, and its
non-negativity is the stochastic analogue of the second law of thermodynamics
\citep{seifert2005entropy,esposito2010three}. For a
\gls{CTMC}, this yields the instantaneous entropy production rate and its time
integral
\begin{align}
  \Eptotdot
  \defeq \int_0^T \Eptot(t)\,dt =
  \int_0^T \sum_{x \neq y}
  p_t(x)\,Q_t(y,x)
  \ln\!\left(
    \frac{p_t(x)\,Q_t(y,x)}{p_t(y)\,Q_t(x,y)}
  \right)\,dt.
  \label{eq:entr_prod}
\end{align}
In continuous diffusion, when the score function is parameterized by a neural network, this quantity is termed \emph{Neural Entropy} \citep{premkumar2024neural},
measuring the total information cost of inverting the corruption process.
For uniform diffusion with strictly positive symmetric rates, the entropy production is finite for any finite horizon $T$ provided $p_t$ has full support. In the context of \glspl{UDM} \cref{eq:entr_prod} can be practically estimated via samples from \(p_t\) and a score network
\begin{align}
  \Eptotdot_\theta
  =
  -\int_0^T \expected_{x\sim p_t}\left[\sum_{y:y\neq x}
  \frac{\sigma(t)}{N-1}
  \ln s_t^\theta(x)_y\right]\,dt.
  \label{eq:entr_prod_uniform}
\end{align}

\subsection{Geometric cost}
The rate at which a generative model can transform an initial noise
distribution $p_T$ into the data distribution $p_0$ is fundamentally
constrained by thermodynamic and geometric considerations. Entropy
production, however, focuses only on information transfer, ignoring
the effective probability mass movement. In stochastic thermodynamics,
this aspect is quantified by the dynamical activity
\citep{van2023thermodynamic}:
\begin{equation}
\mathcal{A}(t) = \sum_{x \neq y} Q_t(x,y)\,p_t(y),
\end{equation}
which counts the total rate of transitions weighted by the occupancy
of the source state. Activity and entropy production jointly constrain the speed of state transformation. This is made precise by the variational formula of
\citet[Corollary~7]{van2023thermodynamic}:
\begin{equation}\label{eq:cor7}
\mathcal{W}(p_0, p_T)
= \min \int_0^T
\frac{\Eptot(t)}{2}\,
\Psi\!\left(\frac{\Eptot(t)}{2\mathcal{A}(t)}\right)^{-1}
dt,
\end{equation}

where $\mathcal{W}$ is the 1-Wasserstein distance, $\Psi$ is the inverse of $x\tanh(x)$ and the minimum is over all admissible Markov jump processes connecting $p_T$ to $p_0$. For a fixed, pre-trained \gls{UDM}, the dynamics are not the minimizer in \cref{eq:cor7}, so the integral evaluated along the \gls{UDM} trajectory upper-bounds $\mathcal{W}(p_0, p_T)$. Moreover, the function $\Psi$ satisfies $\Psi(x) \geq \sqrt{x}$, which yields
\begin{equation}\label{eq:wds_bound}
\frac{\Eptot(t)}{2}\,
\Psi\!\left(\frac{\Eptot(t)}{2\mathcal{A}(t)}\right)^{-1}
\;\leq\;
\sqrt{\frac{\mathcal{A}(t)\,\Eptot(t)}{2}}.
\end{equation}
The geometric mean structure of \cref{eq:wds_bound} balances probability mass displacement
(captured by $\mathcal{A}$) and directed, irreversible motion
(captured by $\Eptot$).

\subsection{Cumulative Progress}
To quantify the progression of the diffusion process, we define the \textit{Cumulative Progress} function $\mathcal{C}(t)=\int_0^t \rho(\tau) \d\tau$, where $\rho(\tau)$ is a non-negative function and represents the instantaneous rate of change of the chosen metric. The objective of our improved scheduling is to find a time discretization $\{t_k\}_{k=0}^K$ such that the progress between consecutive steps is uniform. The improved schedule is obtained by inverting the warping function \(\Phi(t) = \mathcal{C}(t)/\mathcal{C}(T)\). For $K$ generation steps we define the discrete time points $t_k$ as uniform samples in the progress domain:
\begin{equation} \label{eq:inverse_cdf}
    t_k = \Phi^{-1}\left( k/K \right), \quad \text{for } k = 0, \dots, K.
\end{equation}
Next, we cast this framework for our proposed schedules: we define the \gls{EDS} as the schedule obtained by setting \(\rho_{\text{EDS}}(t)\defeq\Eptot_\theta(t)\) and \gls{WDS} as the schedule obtained by setting \(\rho_{\text{WDS}}(t)\defeq\sqrt{\nicefrac{\mathcal{A}}{2}(t)\,\Eptot_\theta(t)}\). Intuitively, our progress functions allow us to allocate equally information production and probability mass movement across sampling steps. \Cref{alg:schedule} highlights the practical estimation pipeline: we use trained models to estimate our rate functions, which only require forward passes, similarly to a training step. Differently from other methods, we do not optimize an objective, resulting in a more efficient schedule estimation.

\section{Experiments}\label{sec:experiments}

In this section, we evaluate our proposed sampling schedules, \gls{EDS} and \gls{WDS} on binomial data, monophonic music and text at different \gls{NFE}. We report more experimental details in \Cref{apdx:exp}.

\begin{figure}[htbp]
    \centering
    
    \begin{subfigure}[b]{0.24\textwidth}
        \centering
        \includegraphics[width=\textwidth]{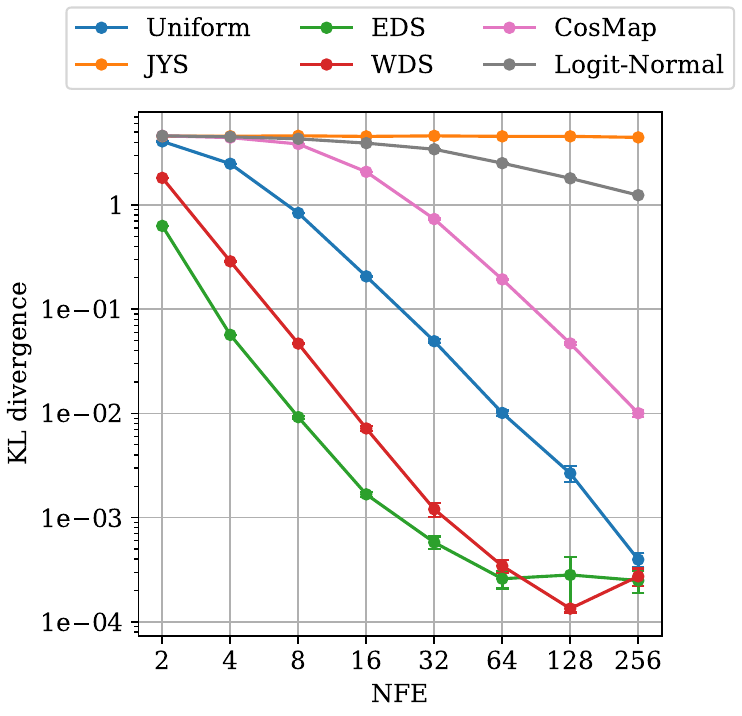}
        \caption{\(n=31\), linear noise schedule}
        \label{fig:top_left} 
    \end{subfigure}
    \hfill
    \begin{subfigure}[b]{0.24\textwidth}
        \centering
        \includegraphics[width=\textwidth]{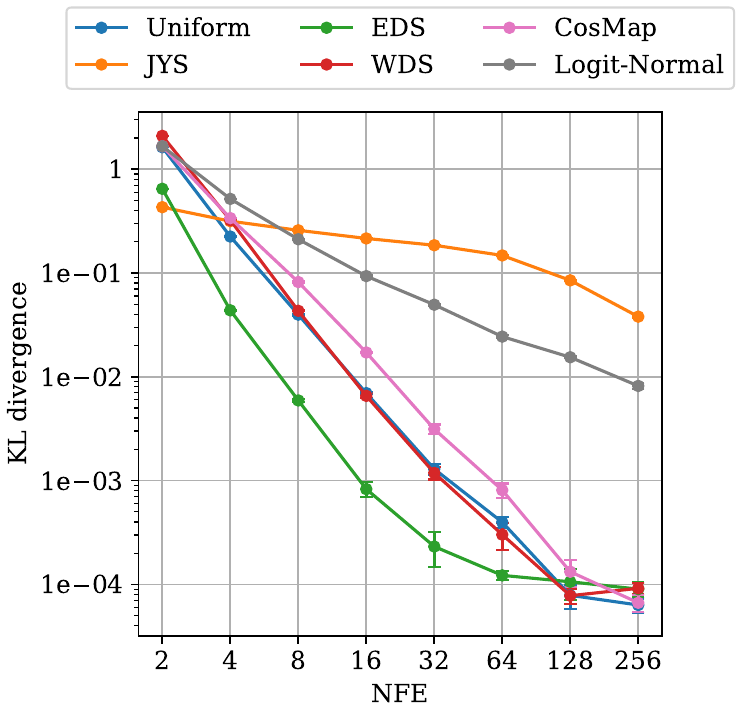}
        \caption{\(n=31\), log-linear noise schedule}
        \label{fig:top_right}
    \end{subfigure}
    \hfill 
    \begin{subfigure}[b]{0.24\textwidth}
        \centering
        \includegraphics[width=\textwidth]{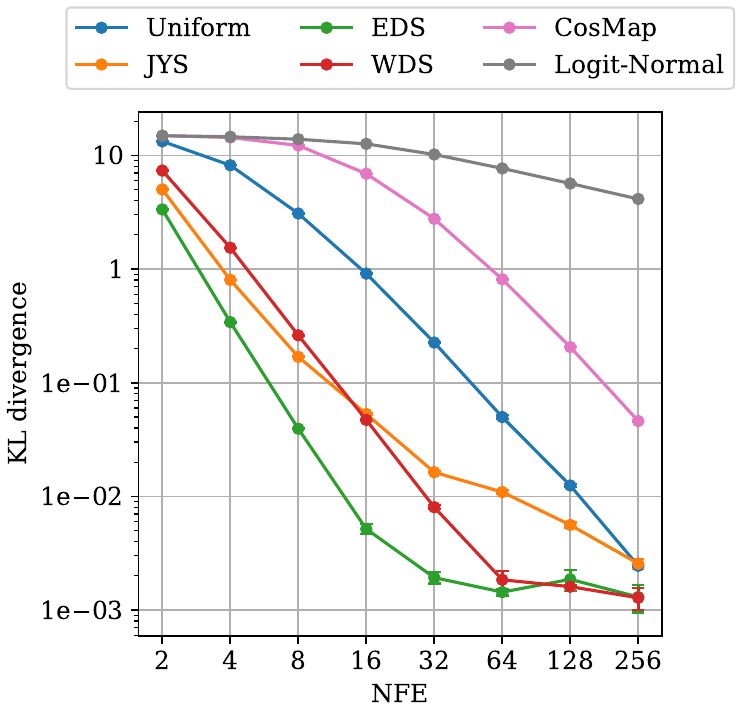}
        \caption{\(n=255\), linear noise schedule}
        \label{fig:bottom_left}
    \end{subfigure}
    \hfill
    \begin{subfigure}[b]{0.24\textwidth}
        \centering
        \includegraphics[width=\textwidth]{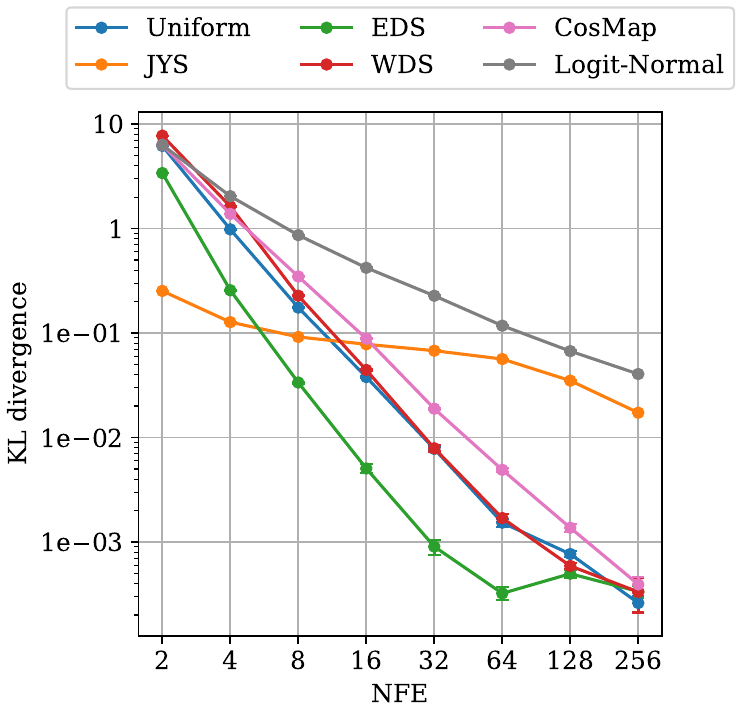}
        \caption{\(n=255\), log-linear noise schedule}
        \label{fig:bottom_right}
    \end{subfigure}
    
    \caption{Binomial distribution benchmark. We report the mean performance across 5 seeds, the bars indicate the standard error of the mean.}
    \label{fig:binomial}
\end{figure}

\begin{figure}[htbp]
    \centering
    
    \begin{subfigure}[b]{0.32\textwidth}
        \centering
        \includegraphics[width=\textwidth]{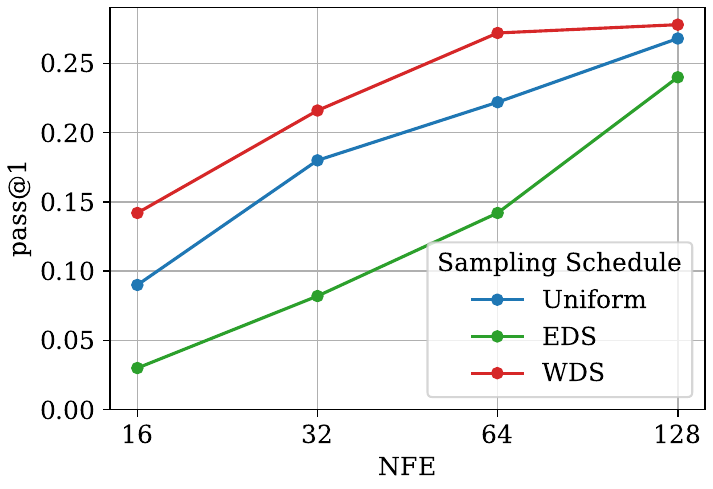}
        \caption{MBPP}
        \label{fig:mbpp} 
    \end{subfigure}
    \hfill
    \begin{subfigure}[b]{0.32\textwidth}
        \centering
        \includegraphics[width=\textwidth]{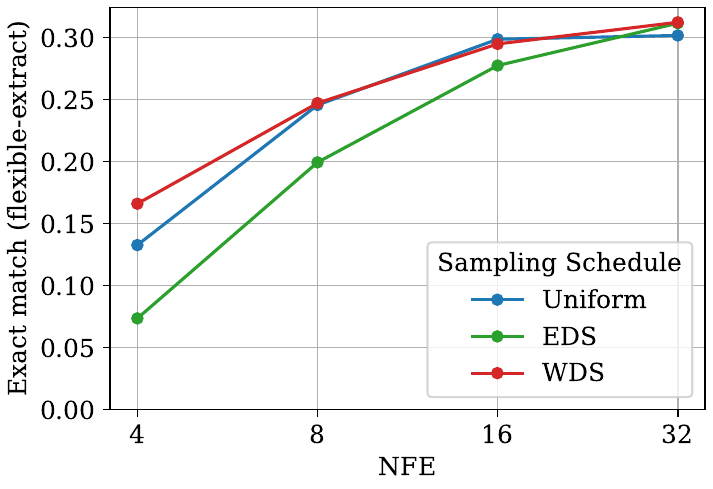}
        \caption{GSM8K}
        \label{fig:gsm8k}
    \end{subfigure}
    \hfill 
    \begin{subfigure}[b]{0.32\textwidth}
        \centering
        \includegraphics[width=\textwidth]{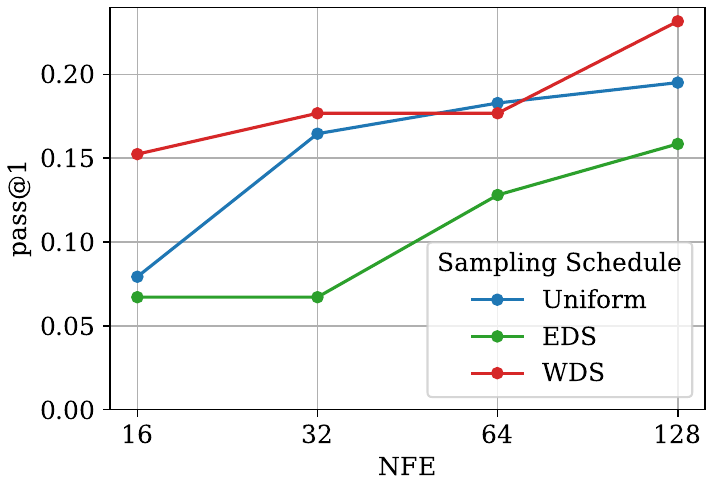}
        \caption{HumanEval}
        \label{fig:humaneval}
    \end{subfigure}
    
    \caption{Sumi reasoning benchmark.}
    \label{fig:sumi}
\end{figure}

\textbf{Binomial Data.} We evaluate unconditional generation of 10-dimensional binomial data with parameters \(n=\{31,255\}\) and \(p=0.5\), using both a discrete diffusion process with log-linear and linear noise schedules, and evaluate the generated samples using the \gls{KL} divergence. In this simplified setting, we have access to the ground truth discrete score, which allows us to isolate the performance of our schedules. We compare against JYS\citep{park2024jump} and the CosMap and Logit-Normal schedules from \citep{esser2024scaling}. \Cref{fig:binomial} shows that both \gls{EDS} and \gls{WDS} outperform other baselines across all settings.

\begin{wrapfigure}{!ht}{0.5\textwidth}
    \centering
    \caption{Performance comparisons on piano note generation across different metrics.}
    \begin{subfigure}{0.48\linewidth}
        \centering
        \includegraphics[width=\linewidth]{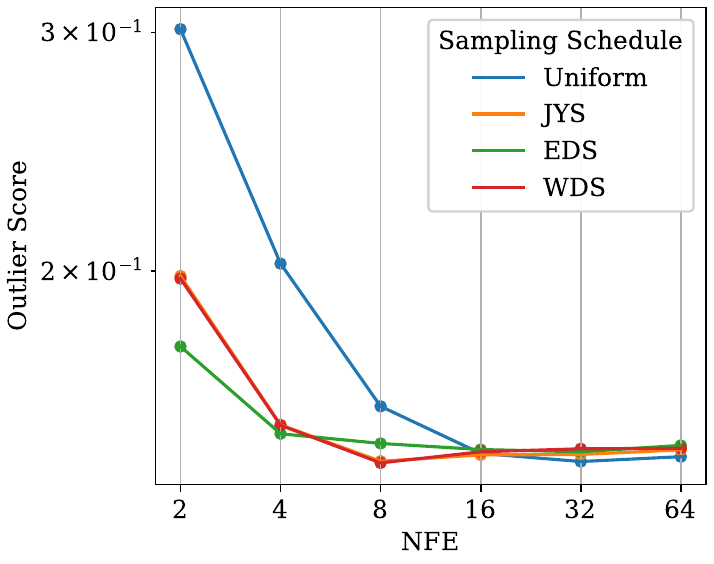}
        \caption{Outlier score.}
        \label{fig:piano-outlier}
    \end{subfigure}\hfill
    \begin{subfigure}{0.48\linewidth}
        \centering
        \includegraphics[width=\linewidth]{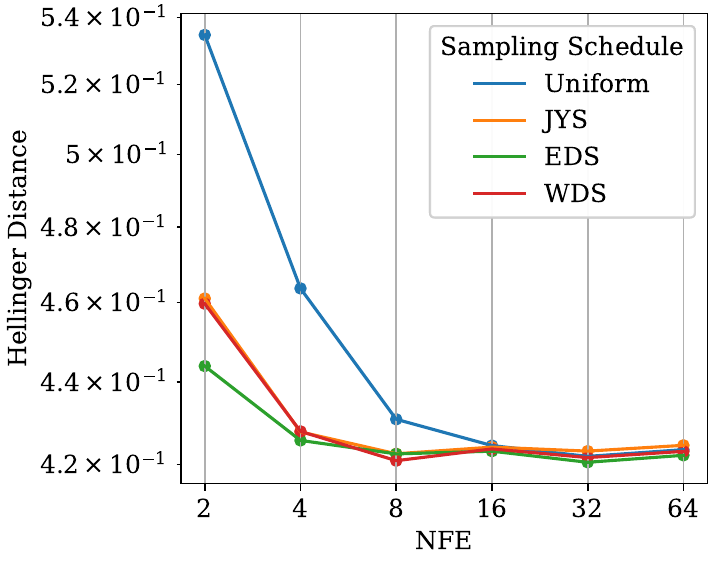}
        \caption{Hellinger distance.}
        \label{fig:piano-hellinger}
    \end{subfigure}

    \label{fig:piano-combined}
\end{wrapfigure}

\textbf{Monophonic Music.} We evaluate conditional music generation on the Lakh pianoroll dataset \citep{raffel2016learning, dong2018musegan}, employing the uniform transition model from \citet{campbell2022continuous}, following the protocol by \citet{park2024jump}. Conditioning on two bars, we generate the subsequent 14 bars (256 total timesteps). Performance is assessed using the outlier score and the Hellinger distance between note distributions at reduced \gls{NFE} ($2$ to $64$) relative to a high-fidelity baseline (\gls{NFE} $512$). Results in \Cref{fig:piano-combined} demonstrate that our proposed schedules, \gls{EDS} and \gls{WDS}, significantly outperform the uniform sampling baseline. Notably, \gls{EDS} matches the quality of $512$-\gls{NFE} samples at a fraction of the cost.

\textbf{Language Modeling.} We evaluate reasoning on zero-shot and few-shot tasks using the large uniform diffusion language model Sumi-7B \citep{ye2026sumi} on Humaneval \citep{humaneval}, GSM8K \citep{gsm8k}, and MBPP \cite{mbpp} benchmarks. We use a generation length of 256 for HumanEval and MBPP and of 64 for GSM8K, following \citep{ye2026sumi}. \Cref{fig:sumi} shows that \gls{WDS} obtains the better performance overall with remarkable gains on low \glspl{NFE} tasks, followed by the uniform schedule, while \gls{EDS} lags behind. Due to the fact that little information is produced in the earlier sampling steps, \gls{EDS} does not take into account the large amount of probability mass that is being moved. As shown in \cref{fig:rate-diagnostics}, \gls{WDS} rates take a very different shape compared to \gls{EDS} due to the dynamical activity, which the latter ignores. We provide numerical results for \gls{WDS} in \cref{tab:sumi_speedup}.
\section{Conclusion} 
In this work, we presented an improved sampling framework that achieves a superior compute-budget performance trade-off. We demonstrated that by applying thermodynamic principles to compute entropy production, we can characterize the evolution of the reverse process. Our analysis established that standard uniform schedules are often inefficient, disproportionately allocating computational steps to regimes with minimal information gain. Motivated by stochastic thermodynamics, we introduced \gls{EDS}, based on entropy production, which aligns to the information dynamics. Building on top of \gls{EDS}, \gls{WDS} balances information cost and dynamical activity, providing substantial gains in the failure modes of \gls{EDS}. Together, \gls{EDS} and \gls{WDS} represent powerful, physics-inspired heuristics, to speed up \gls{UDM}.

\bibliography{sections_arxiv_new/bibliography}
\bibliographystyle{plainnat}

\appendix

\section{Additional details}

\subsection{Implementation details}\label{subsec:impdets}
\Cref{alg:schedule} outlines our method, which is practical and requires no additional training. Using a pre-trained model and a representative batch from the training data, we first estimate the neural entropy rate $\Ena_{\theta}(t)$ across the diffusion trajectory. We then define a cumulative progress function based either on the entropy rate itself (informational progress, \gls{EDS}) or on a Wasserstein distance proxy derived from it (geometric progress, \gls{WDS}) map physical time to an intrinsic domain. Finally, by inverting this function, we obtain a new time schedule that ensures uniform sampling steps within this intrinsic space and is compatible with any sampler (e.g., the Euler \(\tau\)-leaping \cite{campbell2022continuous}).

\begin{algorithm}[!htb]
   \caption{Our Method: Improved Sampling Schedule}
   \label{alg:schedule}
\begin{algorithmic}[1]
   \REQUIRE Pre-trained Model $\theta$, Dataset Batch $\mathbf{X}_0$, Diffusion range $[0, T]$, Steps $K$.
   \ENSURE Discrete schedule $\{t_k\}_{k=0}^K$.
   
   \STATE \textbf{1. Estimate Neural Entropy}
   \STATE Define fine grid $\tau_i \in [0, T]$ (e.g., $N=1000$ points).
   \FOR{each $\tau_i$}
       \STATE Sample batch $x_{\tau_i} \sim \fp_{\tau_i}$ 
       \STATE Estimate $\Eptot_{\theta}(\tau_i)$
   \ENDFOR
   
   \IF{Strategy is EDS}
       \STATE Integrate $\Eptot_{\theta}(\tau)$ to obtain progress curve $\mathcal{C}(\tau)$.
   \ELSIF{Strategy is WDS}
       \STATE Integrate $\sqrt{\nicefrac{\mathcal{A}(t)}{2}\,\Eptot(t)}$ to obtain progress curve $\mathcal{C}(\tau)$.
   \ENDIF
   
   \STATE $\Phi(\tau) \leftarrow \mathcal{C}(\tau) / \mathcal{C}(T)$ \COMMENT{Normalize to $[0, 1]$}
   
   \STATE \textbf{2. Construct Schedule}
   \FOR{$k=0$ \textbf{to} $K$}
       \STATE $t_k \leftarrow \Phi^{-1}(k / K)$ \COMMENT{Uniform sampling in intrinsic space}
   \ENDFOR
   \STATE \textbf{return} $\{t_k\}$
\end{algorithmic}
\end{algorithm}

\subsection{Experiments details}
\label{apdx:exp}
For all experiments, we adhere to the procedure in \Cref{alg:schedule}, which requires evaluating the neural entropy. To achieve this, we use $N$ samples from the training dataset and evaluate them across $N_{\tau}$ timesteps uniformly sampled from $[0,T]$ to estimate the entropy rate. The Wasserstein bound is derived in a similar manner. The derived time schedule is sampler-agnostic. In \Cref{tab:hyperparameters}, we report the parameters $N$ and $N_\tau$ used for our empirical validation.

\begin{table}[h]
    \centering
    \caption{Parameters used for estimating neural entropy across different datasets.}
    \label{tab:hyperparameters}
    \begin{tabular}{lcc}
        \toprule
        \textbf{Dataset} & \textbf{Samples ($N$)} & \textbf{Timesteps ($N_t$)} \\
        \midrule
        Binomial Distributions & 1024 & 1024  \\
        Monophonic Music & 64 & 1024 \\
        Sumi-7B (MBPP) & 64 & 512 \\
        \bottomrule
    \end{tabular}
\end{table}

As our method is training-free, it relies exclusively on pre-trained model checkpoints. In all our experiments, we prioritize using pre-trained models from the official codebase of \citet{park2024jump}; where these are unavailable, we train the models using their provided scripts. We compare our method against JYS \citep{sahoo2024mulan} in binomial and piano note generation experiments. We also provide the schedule computation times for JYS against entropy production at \cref{tab:schedule_comp}. The overhead for interpolating the cumulative progress function for \gls{EDS} and \gls{WDS} is negligible. While the schedule computation times are comparable in small datasets, our methods are much more lightweight in bigger datasets. For Sumi-7B \citep{ye2026sumi}, the neural entropy rate and Wasserstein rate are estimated on the MBPP training split (used only for schedule construction), and the resulting schedule is then evaluated zero-shot on HumanEval \citep{humaneval}, as described in \Cref{sec:experiments}. We report the relative improvements of \gls{WDS} compared to the uniform baseline in reasoning tasks with Sumi-7B at \cref{tab:sumi_speedup}; \gls{WDS} provides the biggest advantages at lower \glspl{NFE}.

\begin{table}[!ht]
    \centering
    \caption{Schedule computation time across different datasets}
    \label{tab:schedule_comp}
    \begin{tabular}{lccc}
\toprule
Dataset & JYS (s) & Entropy production (s) & JYS / Entropy production \\
\midrule
Binomial ($n=255$, linear) & 13.72 & 12.70 & 1.08$\times$ \\
Binomial ($n=255$, loglinear) & 12.37 & 10.91 & 1.13$\times$ \\
Binomial ($n=31$, linear) & 12.24 & 10.51 & 1.16$\times$ \\
Binomial ($n=31$, loglinear) & 12.27 & 11.07 & 1.11$\times$ \\
Piano & 40.65 & 21.66 & 1.88$\times$ \\
\bottomrule
\end{tabular}

\end{table}

\begin{table}[!ht]
    \centering
    \caption{Numerical results for WDS on reasoning tasks}
    \label{tab:sumi_speedup}
    \begin{tabular}{llccc}
\toprule
Task & NFE & Uniform & WDS & Performance Improvement (WDS / Uniform) \\
\midrule
GSM8K & 4 & 0.1327 & 0.1660 & 1.25$\times$ \\
 & 8 & 0.2456 & 0.2472 & 1.01$\times$ \\
 & 16 & 0.2987 & 0.2949 & 0.99$\times$ \\
 & 32 & 0.3017 & 0.3124 & 1.04$\times$ \\
\midrule
MBPP & 16 & 0.0900 & 0.1420 & 1.58$\times$ \\
 & 32 & 0.1800 & 0.2160 & 1.20$\times$ \\
 & 64 & 0.2220 & 0.2720 & 1.23$\times$ \\
 & 128 & 0.2680 & 0.2780 & 1.04$\times$ \\
\midrule
HumanEval & 16 & 0.0793 & 0.1524 & 1.92$\times$ \\
 & 32 & 0.1646 & 0.1768 & 1.07$\times$ \\
 & 64 & 0.1829 & 0.1768 & 0.97$\times$ \\
 & 128 & 0.1951 & 0.2317 & 1.19$\times$ \\
\bottomrule
\end{tabular}

\end{table}

\section{Entropy Production and Wasserstein Rate}
\label{apdx:rates}

\begin{figure}[!ht]
    \centering
    \begin{subfigure}[b]{0.32\textwidth}
        \centering
        \includegraphics[width=\textwidth]{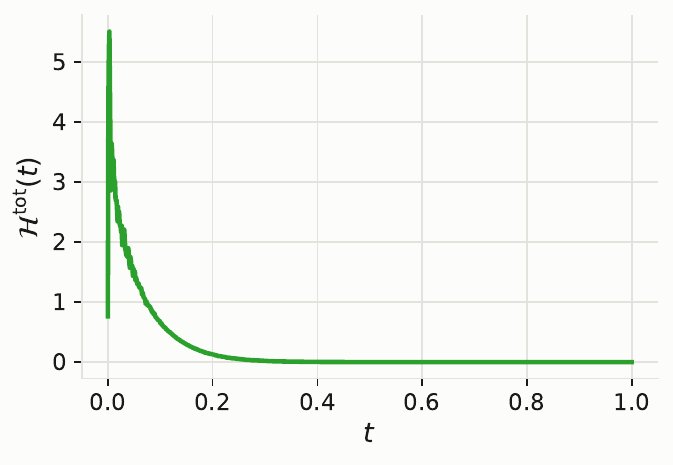}
        \caption{Binomial -- Linear schedule, $\Eptot(t)$}
    \end{subfigure}
    \hfill
    \begin{subfigure}[b]{0.32\textwidth}
        \centering
        \includegraphics[width=\textwidth]{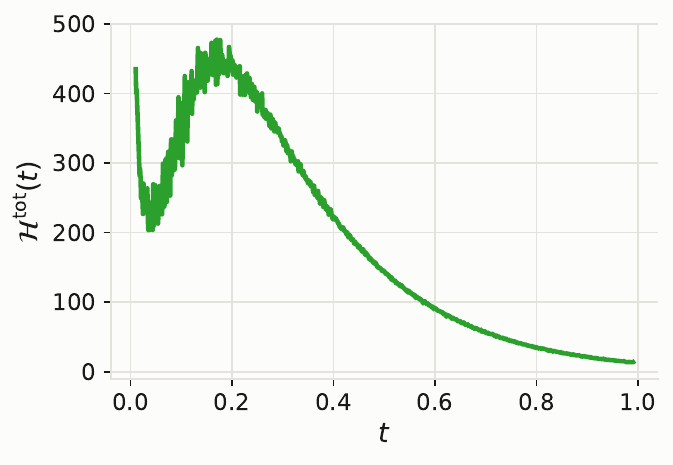}
        \caption{Monophonic Music, $\Eptot(t)$}
    \end{subfigure}
    \hfill
    \begin{subfigure}[b]{0.32\textwidth}
        \centering
        \includegraphics[width=\textwidth]{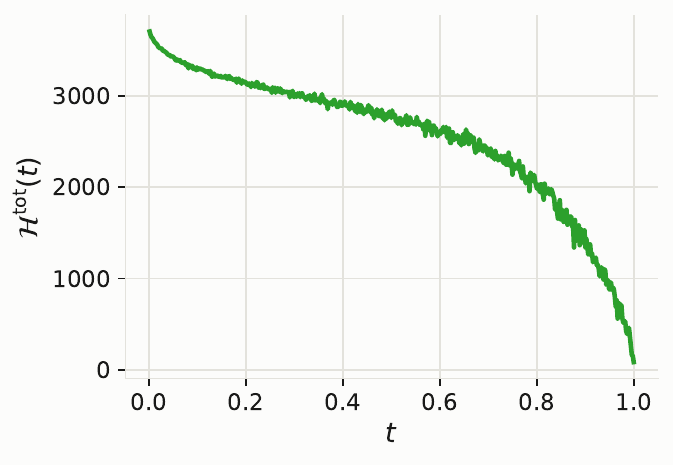}
        \caption{Sumi-7B, $\Eptot(t)$}
    \end{subfigure}

    \begin{subfigure}[b]{0.32\textwidth}
        \centering
        \includegraphics[width=\textwidth]{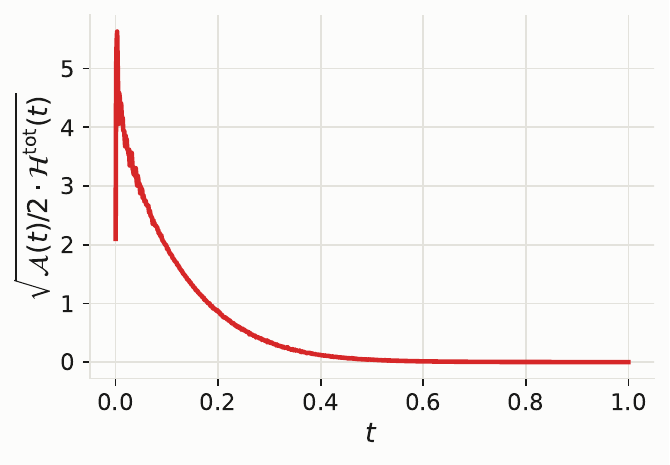}
        \caption{Binomial -- Linear schedule, $\sqrt{\nicefrac{\mathcal{A}(t)}{2}\Eptot(t)}$}
    \end{subfigure}
    \hfill
    \begin{subfigure}[b]{0.32\textwidth}
        \centering
        \includegraphics[width=\textwidth]{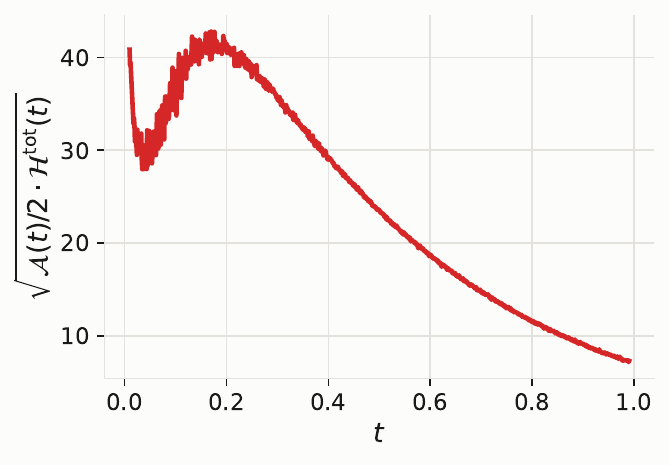}
        \caption{Monophonic Music, $\sqrt{\nicefrac{\mathcal{A}(t)}{2}\Eptot(t)}$}
    \end{subfigure}
    \hfill
    \begin{subfigure}[b]{0.32\textwidth}
        \centering
        \includegraphics[width=\textwidth]{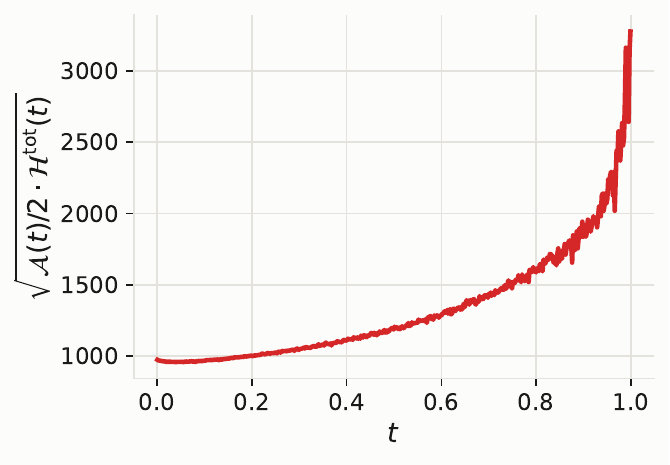}
        \caption{Sumi-7B, $\sqrt{\nicefrac{\mathcal{A}(t)}{2}\Eptot(t)}$}
    \end{subfigure}

    \caption{Entropy production rate (top) and Wasserstein rate (bottom) for Binomial, Monophonic Music, and Sumi-7B.}
    \label{fig:rate-diagnostics}
\end{figure}

\Cref{fig:rate-diagnostics} plots the estimated entropy production rate $\Eptot(t)$ and the Wasserstein rate $\sqrt{\nicefrac{\mathcal{A}(t)}{2}\,\Eptot(t)}$ for Binomial Distribution, Monophonic Music, and Sumi-7B. The two quantities \gls{EDS} and \gls{WDS} respectively equalize across sampling steps (\Cref{eq:inverse_cdf}).

\paragraph{Thermodynamic quantities in specific datasets.} $\Eptot(t)$ is a pure irreversibility measure. The Wasserstein rate additionally weights it by the dynamical activity $\mathcal{A}(t) = \sum_{x\neq y} p_t(x)\,Q_t(x,y)$: the raw rate at which probability mass jumps between states, irrespective of whether that motion is informative. When $\mathcal{A}(t)$ is constant in $t$, as in Monophonic Music experiment, the Wasserstein rate is exactly a square-root-compressed copy of $\Eptot(t)$, so \gls{WDS} closely tracks \gls{EDS}, as visible in \Cref{fig:rate-diagnostics}. When $\mathcal{A}(t)$ instead varies strongly with $t$, as for Sumi-7B, the two rates can be qualitatively different: $\Eptot(t)$ decays monotonically toward $t=1$, while $\sqrt{\nicefrac{\mathcal{A}(t)}{2}\Eptot(t)}$ \emph{grows} toward $t=1$, since the activity's own growth outpaces the entropy rate's decay. This is the key mechanism behind \gls{WDS}'s advantage over \gls{EDS} at low \gls{NFE} on Sumi-7B (\Cref{sec:experiments}): by accounting for activity, \gls{WDS} avoids over-concentrating steps in a regime that looks informative in isolation but where little probability mass is actually moving, allocating more of its budget to the high-activity, high-noise end instead.

\paragraph{Activity formulas for our use cases.} For a \gls{CTMC} with a time-homogeneous, constant off-diagonal rate $c$ over $L$ token positions, such as in Monophonic Music experiments the activity's sum collapses to
\begin{equation}
    \mathcal{A}(t) = L\,c,
\end{equation}
constant in $t$ by construction, which is exactly the flat activity underlying the near-identical shapes in \Cref{fig:rate-diagnostics} for these two experiments. For Sumi-7B, whose forward process resamples each of the $L$ generated tokens uniformly with probability $1-e^{-\cumnoise(t)}$ under the log-linear schedule $\cumnoise(t) = -\ln\!\big(1-t\big)$, the instantaneous noise rate is $\diffusion(t) = \d{\cumnoise}/\d{t} = \tfrac{1}{1-t}$, and the activity is
\begin{equation}
    \mathcal{A}(t) = L\,\diffusion(t) = \frac{L}{1-t},
\end{equation}
monotonically increasing in $t$ and diverging as $t\to1$.

\end{document}